\pdfoutput=1
\documentclass[11pt]{article}
\usepackage[margin=1.1in]{geometry}
\usepackage{amsmath,amssymb,amsthm}
\usepackage{graphicx}
\usepackage{booktabs}
\usepackage{multirow}
\usepackage[numbers,sort&compress]{natbib}
\usepackage[colorlinks=true,linkcolor=blue,citecolor=blue,urlcolor=blue]{hyperref}
\usepackage{xcolor}
\usepackage{tikz}
\usetikzlibrary{arrows,positioning}
\usepackage{authblk}

\newtheorem{theorem}{Theorem}
\newtheorem{lemma}{Lemma}
\newtheorem{proposition}{Proposition}

\theoremstyle{remark}
\newtheorem{remark}{Remark}

\newcommand{\pr}{\mathbb{P}}

\newcommand{\cset}{\mathcal{C}_{1-\alpha}}
\title{ARC: Augmented-Rank Conformalization for Changepoint Localization
--- Finite-Sample Validity and Distribution-Robust Efficiency}

\author[1,2]{Chenchen Peng\thanks{Corresponding author. Email:
\texttt{pengcc@emails.bjut.edu.cn}}}
\author[1]{Mixia Wu}
\author[1]{Qijing Yan}
\author[2]{Zhiqi Shen}
\author[2]{Jie Zhang}

\affil[1]{School of Mathematics, Statistics and Mechanics, Beijing
University of Technology, Beijing 100124, China}
\affil[2]{College of Computing and Data Science, Nanyang Technological
University, Singapore 639798, Singapore}

\date{August 7, 2026}

\begin{document}
\maketitle

\begin{abstract}
Conformal changepoint localization converts any score function into a
confidence set for the changepoint location with finite-sample coverage.
In this precise sense coverage has become universal, since it is
guaranteed irrespective of how the score is chosen; efficiency has not.
The oracle score is a likelihood ratio, so practical scores estimate
density ratios, and the length of the resulting sets deteriorates
silently under heavy tails, skewness, and distribution shift, precisely
the regimes in which no existing length guarantee applies. We propose
ARC (Augmented-Rank Conformalization), a family of scores that depend
on the data only through within-segment ranks: rank-CUSUM location and
scale channels, their fixed combinations, and a lightweight neural
score trained on synthetic data and then frozen. Every ARC score
inherits finite-sample coverage for every frozen weight configuration,
including random initializations and misspecified training. Our main
result is an efficiency transfer theorem. The entire ARC confidence set
is almost surely invariant under strictly increasing marginal
transforms, so the law of the set length depends on the data pair only
through its rank structure: lengths certified once hold verbatim across
the pair's entire monotone orbit, however heavy-tailed or skewed,
whereas a plug-in score's length changes with every re-expression.
Across different rank structures lengths do change, and are reported as
such. Classical rank-test theory positions ARC scores as targeting the
optimal invariant score, with a bounded price of invariance.
Simulations confirm nominal coverage for all scores, including
deliberately sabotaged networks, exactly identical sets under monotone
transforms where plug-in likelihood scores inflate, and set lengths
that deteriorate smoothly where plug-in sets become vacuous; on the
well-log benchmark ARC localizes annotated shifts to three to five
candidates and signals model misfit through an empty set. Two
boundaries are charted rather than hidden: serial dependence destroys
exactness and block permutations repair it only partly, and trend-type
alternatives lie outside the piecewise-exchangeable model.
\end{abstract}


\section{Introduction}\label{sec:intro}

Practitioners who run a changepoint detector on sensor streams, activity
recordings, or monitoring data are typically supplied with a point
estimate $\hat\tau$ and nothing further. The decisions that follow,
concerning when to retrain a model, whether to trust a segment
boundary, and how much data to discard, depend on the uncertainty
attached to that location, which is rarely quantified
\citep{li2026modern,steland2026online,zhang2026changeaware}. In the
real-data application considered below, the classic well-log drilling
series \citep{fearnhead2019changepoint}, the analysis windows comprise
a few dozen observations around a suspected shift, the measurements are
contaminated by outlier bursts, and the marginal distribution is
unknown, so any uncertainty statement resting on a parametric model is
fragile at the outset.

A recent line of work has solved one half of this problem. Conformal
changepoint localization \citep[rooted in conformal
prediction,][]{vovk2005algorithmic} constructs, from an arbitrary score
function, a
set $\cset\subseteq\{1,\dots,n\}$ of candidate locations with the
finite-sample guarantee $\pr(\tau\in\cset)\ge 1-\alpha$ under no
distributional assumption beyond within-segment exchangeability: the
matrix-of-p-values construction of \citet{dandapanthula2026offline}, the
CONCH framework of \citet{hore2026conformal}, and, in the sequential
setting, post-detection localization
\citep{saha2026post,saha2026distributionfree}. The defining feature of
these results is universality: an arbitrary score yields valid
coverage, and \citet{hore2026conformal} prove that every valid
distribution-free localization procedure arises in this manner.

The score question has been answered only at the oracle. Both
\citet{dandapanthula2026offline} and \citet{hore2026conformal} establish
conformal Neyman--Pearson lemmas: the score minimizing the expected set
length is a likelihood ratio between the pre- and post-change
distributions. In practice these densities are unknown, so the recommended
implementations estimate the ratio with pre-trained classifiers or plug-in
models, and the available length guarantees
\citep{hore2026conformal,bhattacharya2025length} assume the oracle ratio
or a consistent estimate of it. No existing result addresses the length
of the sets when the plug-in is misspecified, when the tails are heavy,
or when the deployment distribution differs from the one on which the
score was tuned, which are precisely the regimes that motivated
distribution-free coverage in the first place.

This is the gap addressed here. Conformal wrappers rendered coverage
universal, but no current score offers transferable efficiency. Set
lengths certified under one distribution carry no guarantee under
another, and the oracle likelihood-ratio score, being density-dependent,
is the score whose efficiency transfers worst. A score family whose
entire localization output is invariant to the marginal distribution of
the data has not previously been considered.

We propose ARC (Augmented-Rank Conformalization). The score family
measures within-segment changepoint evidence through within-segment
ranks alone, comprising rank-CUSUM location and scale channels, their
fixed maxima or simplex combinations, and a single-hidden-layer network
that reads downsampled rank paths, is trained on synthetic data, and is
frozen before deployment. Supplied to the backend of
\citet{hore2026conformal} or \citet{dandapanthula2026offline}, every
ARC score yields finite-sample coverage for every frozen weight
configuration, including random initializations and misspecified
training (Theorem~\ref{thm:coverage}).

The main theoretical contribution is an efficiency transfer theorem.
Because ranks are invariant under strictly increasing transforms, the
entire ARC confidence set is almost surely unchanged when an arbitrary
strictly increasing $g$ is applied to the data
(Theorem~\ref{thm:invariance}), and consequently the law of the set
length depends on $(F_0,F_1)$ only through its rank structure. Lengths
certified once transfer verbatim across the entire monotone orbit of
the certified pair, including arbitrarily heavy-tailed or skewed
re-expressions of the same rank experiment, whereas a plug-in score's
length changes with every re-expression. Between genuinely different
rank experiments, such as a Gaussian shift and a Cauchy shift, lengths
differ for every method, and Section~\ref{sec:sim} reports those
differences as they stand. Classical theory
\citep{hajek1967theory,chernoff1958asymptotic} positions ARC as
targeting the optimal invariant score, with a bounded price of
invariance.

Localization is additionally type-aware: location and scale channels
are combined by rules frozen before the test window is seen, so type
adaptivity costs nothing in validity, and the residual gap to an oracle
that knows the change type is quantified empirically. Every number in
Section~\ref{sec:sim} derives from executed simulations, covering
nominal coverage for all scores including deliberately sabotaged
networks, exactly identical sets under monotone transforms where plug-in
likelihood scores drift and inflate, stable lengths across
distributions, and a boundary study under AR(1) dependence in which
block permutations restore coverage only approximately. Trend-type
alternatives are excluded by the model itself
(Remark~\ref{rem:scope}).

Section~\ref{sec:background} places these contributions in the landscape
of localization guarantees. Section~\ref{sec:method} defines the ARC
pipeline, Section~\ref{sec:theory} states the theory,
Section~\ref{sec:sim} reports simulations, Section~\ref{sec:real}
specifies the real-data protocol, and Section~\ref{sec:discussion}
discusses limitations.

\section{Background: Three Guarantees and One Gap}\label{sec:background}

\subsection{Three kinds of localization guarantee}\label{sec:guarantees}

Statements about ``where the changepoint is'' come in three distinct
mathematical forms, which the literature does not always keep apart.

\begin{table}[t]
\centering
\caption{Three localization guarantees. ARC occupies the third row; the
three are complementary, not competing.}
\label{tab:guarantees}
\small
\begin{tabular}{p{3.4cm}p{4.6cm}p{5.6cm}}
\toprule
Guarantee & Form & Representatives \\
\midrule
Region false-positive control &
$\pr\{\exists\,\hat R\in\hat{\mathcal R}:\hat R\cap T^*=\emptyset\}\le\alpha$ &
NSP \citep{fryzlewicz2024nsp}, NOT \citep{baranowski2019not}, SMUCE
\citep{frick2014multiscale}, confidence regions of
\citet{fang2020segmentation}, ART localization
\citep[Thm.~4, Prop.~1]{cui2026art} \\
\addlinespace
Post-detection error control &
FWER over detected $\hat\tau_j$'s; coverage conditional on a correct alarm &
TUNE \citep{jia2024tune}, ART diagnostic \citep[Thm.~5]{cui2026art},
selective inference \citep{jewell2022post}, sequential post-detection
\citep{saha2026post,saha2026distributionfree} \\
\addlinespace
Coverage of the location &
$\pr(\tau\in\cset)\ge 1-\alpha$ &
MCP \citep[Thm.~4.1]{dandapanthula2026offline}, CONCH
\citep[Thm.~3.1]{hore2026conformal}; \textbf{this paper} (score design) \\
\bottomrule
\end{tabular}
\end{table}

The first form controls the probability that a \emph{reported region}
fails to contain a true changepoint; under weak signal the procedure may
return no region at all, which is vacuously correct but uninformative
about where $\tau$ is. The second form diagnoses changepoints that some
detector has already produced. The third form is a genuine confidence set
for the location: it is never empty of meaning, since weak signal
manifests as a longer set rather than silence. ART \citep{cui2026art}
provides sophisticated instances of the first two forms in a
distribution-free way, but not the third; the conformal line
\citep{dandapanthula2026offline,hore2026conformal} provides the third for
an arbitrary score. Our contribution is entirely on the score side of the
third row, and an ARC pipeline can be run alongside an ART diagnostic
without conflict.

\subsection{What is known about the score}\label{sec:scoreknown}

For the third guarantee the wrapper is settled and the oracle is known.
\citet{dandapanthula2026offline} prove a conformal Neyman--Pearson lemma:
the uniformly most powerful conformal score is the likelihood ratio
$q/r$ of post- to pre-change densities; \citet{hore2026conformal} derive
the analogous optimal changepoint-profile score and show
(their Theorem~4.3) that any strictly increasing transformation \emph{of
the score} leaves the confidence set unchanged. Length guarantees exist
under the oracle score or a ratio estimated consistently on independent
data \citep[Thms.~5.1--5.2]{hore2026conformal}, and in parametric
triangular arrays \citep{bhattacharya2025length}. The universality theorem
of \citet[Thm.~6.1]{hore2026conformal} shows every valid procedure is a
CONCH instance for some score; ARC is therefore, by construction, a
concrete score family inside their framework, and we claim no novelty in
the wrapper.

What is missing is any statement about score behaviour \emph{away} from
the oracle. Estimating a density ratio is exactly the kind of task that
heavy tails and distribution shift make hard
\citep{li2026robust,tang2026online}; a classifier tuned on Gaussian
training data retains coverage everywhere (universality) but its set
lengths are unprotected. No existing result bounds, transfers, or even
measures this degradation. The rank-test literature
\citep{hajek1967theory,chernoff1958asymptotic,wilcoxon1945individual,%
mood1954asymptotic} has studied invariant alternatives to
likelihood-based statistics for seventy years, and recent detectors feed
such fixed transforms into calibrated downstream pipelines
\citep{nie2026persistent}, while fully learned critics
\citep{li2026cusumnet} act on raw values and enjoy no invariance. None of
these lines has been connected to conformal localization. Establishing
that connection is the contribution of this paper.

\section{Methodology: The ARC Pipeline}\label{sec:method}

\subsection{Setting and the conformal backend}\label{sec:setting}

We observe a window $X=(X_1,\dots,X_n)$ of real-valued observations with
a single changepoint $\tau$: $X_1,\dots,X_\tau$ i.i.d.\ $F_0$ and
$X_{\tau+1},\dots,X_n$ i.i.d.\ $F_1$, with $F_0\neq F_1$ continuous.
Candidates range over $\mathcal T=\{m_0,\dots,n-m_0\}$ for a margin
$m_0$. For a candidate $t$, the split hypothesis is
\[
H_{0,t}:\ X_1,\dots,X_t \text{ i.i.d.\ and } X_{t+1},\dots,X_n
\text{ i.i.d.},
\]
which holds exactly when $t=\tau$. Fix a \emph{segment statistic} $A$
mapping a segment to a scalar measure of internal changepoint evidence,
and define the split statistic
$W_t=\max\{A(X_{1:t}),\,A(X_{t+1:n})\}$: if $t\neq\tau$, one of the two
segments straddles the true changepoint and $W_t$ tends to be large.
The p-value for $H_{0,t}$ is computed by within-segment permutation.
Drawing $B$ i.i.d.\ pairs of independent uniform permutations
$(\pi_\ell^{(b)},\pi_r^{(b)})$ of the two segments and writing
$W_t^{(b)}$ for the statistic on the permuted window,
\begin{equation}\label{eq:pval}
p_t \;=\; \frac{1+\#\{b: W_t^{(b)}\ge W_t\}}{B+1},
\qquad
\cset \;=\; \{t\in\mathcal T: p_t>\alpha\}.
\end{equation}
Under $H_{0,t}$ both segments are separately exchangeable, so $p_t$ is a
valid Monte Carlo permutation p-value for \emph{any} statistic $A$
\citep{hemerik2018exact,lehmann2005testing}, and
$\pr(\tau\in\cset)=\pr(p_\tau>\alpha)\ge 1-\alpha$. This is the
within-segment instance of the conformal localization schema
\citep{hore2026conformal,dandapanthula2026offline}; everything that
follows concerns the choice of $A$.

\subsection{The ARC score family}\label{sec:family}

Let $R=(R_1,\dots,R_m)$ denote the ranks of a segment of length $m$ (no
ties, almost surely) and $C_i=R_i-(m+1)/2$ the centred ranks. ARC
statistics are functionals of $R$ alone.

\paragraph{Location channel.} The two-sample Wilcoxon evidence for an
internal split at $s$ \citep{wilcoxon1945individual},
\[
Z^{\mathrm{loc}}_s=\Bigl|\sum_{i\le s}C_i\Bigr|\Big/
\sqrt{s(m-s)(m+1)/12},
\qquad
A^{\mathrm{loc}}(R)=\max_{m_1\le s\le m-m_1}Z^{\mathrm{loc}}_s ,
\]
with an interior margin $m_1$; the maximization over $s$ is the
CUSUM-scan device of \citet{page1954continuous} applied to rank scores.

\paragraph{Scale channel.} Mood-type squared centred ranks
\citep{mood1954asymptotic}: with $q_i=C_i^2$ and $\bar q$, $v_q$ the
segment mean and variance of $(q_i)$ (constants across permutations,
since the rank multiset is fixed),
\[
Z^{\mathrm{sc}}_s=\Bigl|\sum_{i\le s}(q_i-\bar q)\Bigr|\Big/
\sqrt{s(m-s)v_q/m},
\qquad
A^{\mathrm{sc}}(R)=\max_{m_1\le s\le m-m_1}Z^{\mathrm{sc}}_s .
\]

\paragraph{Fixed combinations.} The \emph{combined} score is
$A^{\mathrm{comb}}=\max\{A^{\mathrm{loc}},A^{\mathrm{sc}}\}$; a
\emph{learned} convex combination
$A^{w}=w_1A^{\mathrm{loc}}+w_2A^{\mathrm{sc}}$, $w_1+w_2=1$,
$w_1,w_2\ge0$, uses weights fitted on synthetic training data and
frozen thereafter.

\paragraph{Frozen network score.} The two standardized rank paths
$(Z_s^{\cdot})_s$ are downsampled to eight points each, giving a
16-dimensional feature $\phi(R)$, and a single-hidden-layer network
$g_\theta$ ($16\to 64\to 1$, $\tanh$) is trained with Adam
\citep{kingma2015adam} on synthetic Gaussian segments labelled by the
presence of an internal changepoint, then \emph{frozen}:
$A^{\mathrm{nn}}(R)=g_\theta(\phi(R))$. Training data are synthetic and
independent of any test window, so the score is fitted once and then
deployed unchanged.

\begin{figure}[t]
\centering
\resizebox{\textwidth}{!}{%
\begin{tikzpicture}[
  font=\footnotesize,
  box/.style={draw, rounded corners=2pt, align=center, inner sep=4pt,
              minimum height=8mm},
  chan/.style={box, fill=blue!8, minimum width=30mm},
  frozen/.style={box, fill=orange!12},
  outp/.style={box, fill=green!10},
  lab/.style={font=\scriptsize\itshape, text=black!60, align=center},
  arr/.style={-latex, thick, black!70},
  node distance=4mm and 9mm]
\node[box] (win) {test window\\$X_1,\dots,X_n$};
\node[box, right=of win] (rank) {within-segment\\ranks $R$ at each\\candidate split $t$};
\node[chan, right=of rank, yshift=13mm] (loc) {location channel\\rank CUSUM $A^{\mathrm{loc}}$};
\node[chan, right=of rank] (sc) {scale channel\\squared-rank $A^{\mathrm{sc}}$};
\node[chan, right=of rank, yshift=-13mm] (nn) {network score $A^{\mathrm{nn}}$\\$g_\theta(\phi(R))$, $16{\to}64{\to}1$};
\node[frozen, right=of sc, xshift=2mm] (combi) {fixed combination\\$\max$ or $w_1,w_2$\\(frozen before test)};
\node[box, right=of combi] (perm) {within-segment\\permutation\\p-value $p_t$};
\node[outp, right=of perm] (out) {existence test\\change type\\$\hat\tau=\arg\max_t p_t$\\$\cset=\{t: p_t>\alpha\}$};
\node[lab, below=1mm of nn] {trained on synthetic data, then frozen:\\any weight state keeps coverage (Thm.~1)};
\node[lab, above=1mm of loc] {ranks only $\Rightarrow$ monotone-invariant (Thm.~2)};
\draw[arr] (win) -- (rank);
\draw[arr] (rank) -- (loc);
\draw[arr] (rank) -- (sc);
\draw[arr] (rank) -- (nn);
\draw[arr] (loc) -- (combi);
\draw[arr] (sc) -- (combi);
\draw[arr] (nn) -- (combi);
\draw[arr] (combi) -- (perm);
\draw[arr] (perm) -- (out);
\end{tikzpicture}}
\caption{The ARC mechanism. Each candidate split $t$ is scored through
within-segment ranks only; scores are combined by a rule frozen before
the test window is seen and calibrated by exact within-segment
permutation, yielding the localization confidence set together with the
existence test and change-type label.}
\label{fig:pipeline}
\end{figure}
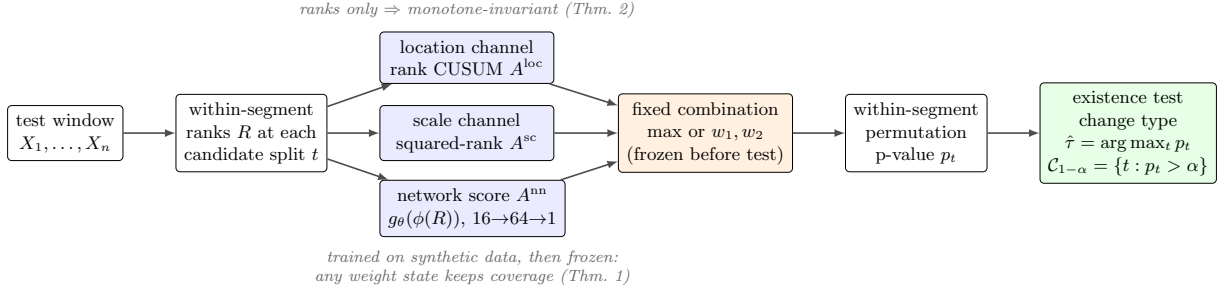

Every member above is a deterministic function of the within-segment
ranks. That single structural property drives the entire theory.
Figure~\ref{fig:pipeline} summarizes the mechanism.

\subsection{Type-aware combination without paying for it}
\label{sec:typeaware}

One might first classify the change type on the test window and then
select the matching channel. Such a rule makes the score a function of
the test data and invalidates the permutation argument behind
\eqref{eq:pval}. ARC therefore fixes the combination rule, either the
max-combination or simplex weights learned on synthetic data, before
the test window is seen, so that validity is retained at no cost,
and we quantify the residual gap to an oracle that knows the change type
in Section~\ref{sec:sim-ablation}. Adaptive selection on the test window
is possible in principle via sample splitting, at a direct cost in
effective window length; we do not pursue it here.

\subsection{The full pipeline and reporting convention}
\label{sec:pipeline}

On a test window ARC reports, in order: (i) an exact existence test,
namely the ART test of \citet[Thms.~1,\,3]{cui2026art} or, equivalently, the
permutation p-value of the combined score on the whole window; (ii) a
change-type label from a type classifier trained on synthetic data and
frozen before deployment; (iii) the point estimate
$\hat\tau=\arg\max_t p_t$; and (iv) the confidence set $\cset$.
The set is reported \emph{unconditionally}: its coverage
$\pr(\tau\in\cset)\ge1-\alpha$ is marginal and does not condition on the
existence test rejecting. Conditional-on-detection coverage is a
strictly stronger target that requires the machinery of
\citet{saha2026post,saha2026distributionfree}; we flag the distinction
explicitly to avoid over-claiming.

\subsection{Weak dependence: block variant}\label{sec:block}

Within-segment exchangeability fails under serial dependence. Replacing
uniform permutations in \eqref{eq:pval} by within-segment circular block
permutations \citep[in the spirit of][]{kunsch1989jackknife} restores
approximate validity, at block length $L$ chosen to exceed the dependence
range. As a practical rule we take $L$ to be two to three times the
smallest lag at which the sample autocorrelation of the segments becomes
negligible; Appendix~\ref{app:ar1} uses $L=10$ for AR(1) with $\phi=0.5$,
whose autocorrelation falls below $0.1$ by lag~$3$.
Coverage is then approximate, not exact; Appendix~\ref{app:ar1} quantifies the
degradation and the repair under an AR(1) design. All Gaussian-through-skewed
claims of exactness in this paper are for independent-within-segment data.

\section{Theory}\label{sec:theory}

\begin{theorem}[Finite-sample coverage, any frozen score]
\label{thm:coverage}
Let $A$ be any ARC score, including $A^{\mathrm{nn}}$ with an arbitrary
frozen parameter $\theta$ obtained by training to convergence, by
stopping after one epoch, by random initialization, or by training on a
misspecified or adversarially labelled synthetic distribution. If
$\theta$ and the
combination rule are fixed independently of the test window, then for
any $\alpha\in(0,1)$ and any $B\ge 1$,
$\pr(\tau\in\cset)\ge 1-\alpha$.
\end{theorem}

The proof (Appendix~A) is the within-segment permutation argument;
validity is inherited from the backend schema of
\citet{hore2026conformal,dandapanthula2026offline} and the exactness of
Monte Carlo permutation tests \citep{hemerik2018exact}. The content is
the quantifier: \emph{no} property of the training run is assumed. The
practical consequence is that ARC needs no deployment-side calibration
data and cannot be invalidated by a bad training day.

\begin{remark}[Model scope]\label{rem:scope}
Theorem~\ref{thm:coverage} is a statement about the piecewise-i.i.d.\
or more generally piecewise-exchangeable, model, which covers arbitrary
distributional changes in location, scale, shape, or any combination.
Trend-type alternatives, in which a deterministic drift follows the
change, place a non-exchangeable segment on one side of the true
split, so $H_{0,\tau}$ itself fails and no procedure in the conformal
localization schema \citep{dandapanthula2026offline,hore2026conformal}
guarantees coverage there, whatever the score. Such alternatives are
outside the scope of this paper, and localizing them with rigorous
uncertainty statements is open.
\end{remark}

\begin{lemma}[Rank invariance]\label{lem:ranks}
Let $g:\mathbb R\to\mathbb R$ be strictly increasing. For any segment,
the rank vector of $(g(x_i))_i$ equals that of $(x_i)_i$; hence every ARC
score, every split statistic $W_t$, and, under the same permutation
randomness, every p-value $p_t$ computed from $g(X)$ coincides with its
value computed from $X$.
\end{lemma}

\begin{theorem}[Invariance and efficiency transfer]\label{thm:invariance}
Couple the procedure on $X$ and on $g(X)$ by sharing the Monte Carlo
permutations. Then $\cset(g(X))=\cset(X)$ almost surely, for every
strictly increasing $g$. Consequently, if $X$ follows the changepoint
model $(F_0,F_1,\tau)$ and $g(X)$ the model
$(F_0\circ g^{-1},F_1\circ g^{-1},\tau)$, the joint law of
$\bigl(\cset,\tau\bigr)$, and in particular the coverage, the
distribution of the set length $|\cset|$, and the localization error of
$\hat\tau$, is identical across the entire orbit
$\{(F_0\circ g^{-1},F_1\circ g^{-1}): g \text{ strictly increasing}\}$.
\end{theorem}

Theorem~\ref{thm:invariance} is the efficiency transfer result: a length
distribution certified once, for instance by Gaussian simulation, holds
verbatim for every monotone-equivalent pair, whether lognormal,
logistic-compressed, or arbitrarily heavy-tailed images of the same
rank configuration. Two remarks delimit the claim.

\begin{remark}[Relation to score-level invariance]
Theorem~4.3 of \citet{hore2026conformal} states that strictly increasing
transformations of the score do not change the CONCH set, a
property of the wrapper that holds for every score. Theorem~\ref{thm:invariance}
concerns strictly increasing transformations \emph{of the data}, and holds
for rank-based scores only: a plug-in likelihood or classifier score
changes with $g$, and its set with it. The two statements are orthogonal,
and only the second yields efficiency transfer.
\end{remark}

\begin{remark}[The price of invariance]\label{rem:price}
The monotone group's maximal invariant is the rank vector
\citep{hajek1967theory}, so every invariant procedure is a rank
procedure, and the optimal invariant score is a rank functional. Against
the oracle likelihood-ratio score of
\citet{dandapanthula2026offline,hore2026conformal}, ARC pays the
classical price of invariance, which rank-test theory bounds tightly: in
the two-sample location problem, normal-scores rank statistics attain
asymptotic relative efficiency at least one against the $t$-statistic
uniformly over distributions \citep{chernoff1958asymptotic}, and the
Wilcoxon channel's ARE at the Gaussian is $3/\pi\approx0.955$. ARC
trades at most a few percent of efficiency at any fixed model for exact
efficiency transfer across the model's entire monotone orbit.
\end{remark}

\begin{proposition}[Length contraction]\label{prop:length}
Consider the location channel with a stochastically ordered alternative,
$\rho=\pr(X'>X)\neq 1/2$ for $X\sim F_0$, $X'\sim F_1$, and
$\tau=\lfloor\gamma n\rfloor$, $\gamma\in(0,1)$ fixed. Then
$|\cset|/n\to 0$ in probability as $n\to\infty$, with the exclusion of
any fixed fraction $|t-\tau|\ge\varepsilon n$ occurring at exponential
rate in $n$.
\end{proposition}

The proof sketch (Appendix~A) combines two-sample $U$-statistic
concentration for the Wilcoxon evidence inside the mislocated segment
with the $O(\sqrt{\log n})$ growth of permutation quantiles; the
argument parallels the length analyses of
\citet{hore2026conformal,bhattacharya2025length} and the interval-width
rates of \citet[Prop.~1]{cui2026art} and \citet{madrid2021optimal}. We
state it as a proposition rather than a sharp-rate theorem: the constants
are not optimized, and our evidence for finite-sample length behaviour is
the simulation study, reported in full. The proposition's scope should
be read precisely: it covers the location channel under a stochastically
ordered alternative ($\rho\neq1/2$); a pure scale change may have
$\rho=1/2$, and for the combined, learned, and network scores we state
no length guarantee; their finite-sample length behaviour is documented
empirically, without theoretical cover, in Section~\ref{sec:sim}.

\section{Simulation Studies}\label{sec:sim}

All results in this section were produced by executing the accompanying
code (files \texttt{arc\_core.py} and \texttt{run\_experiments.py} in
the supplement); every number below is read from the generated result
files. Window length
$n=120$, changepoint $\tau=60$, candidate margin $m_0=10$, interior
margin $m_1=5$, $B=79$ permutations, nominal level $\alpha=0.10$
throughout; Monte Carlo standard errors for coverage entries at 250--300
replications are about $0.02$, and each table states its per-cell
replication count. With $B=79$ the p-values in \eqref{eq:pval} lie on
the grid $\{1/80,\dots,80/80\}$, so the rule $p_t>\alpha$ at
$\alpha=0.10$ retains candidates with $p_t\ge 9/80$ and the procedure
is conservative by at most one grid step;
Appendix~\ref{app:bsens} verifies that coverage and mean length are
insensitive to $B\in\{79,199,999\}$. The network score is trained once on
6{,}000 synthetic Gaussian segments with mean- and scale-type internal
changes in equal proportion and frozen; training uses the
binary-cross-entropy loss with Adam (learning rate $10^{-3}$, batch
size $64$, $30$ epochs; weights initialized $N(0,1/\sqrt{d_{\mathrm{in}}})$;
training AUC $0.986$). Four
weight states are carried through all experiments: \emph{trained}
(30 epochs), \emph{epoch-1}, \emph{random} (never trained), and
\emph{mistrained} (30 epochs on permuted labels). The learned convex
combination, fitted on the same synthetic data by projected gradient
descent on the logistic loss (learning rate $0.05$, $200$ epochs), is
$w=(0.51,\,0.49)$ over (location, scale), mirroring the balanced
composition of the training alternatives. Throughout, ``plug-in LR''
labels the two-sample $t$ statistic, which is a monotone transformation
of the Gaussian likelihood-ratio statistic; it stands for the
density-based plug-in family, not for a universal likelihood-ratio
estimator.

\subsection{Coverage: any frozen weight}\label{sec:sim-coverage}

Table~\ref{tab:coverage} reports empirical coverage under a mean shift
($\delta=2$) for four marginal distributions and nine scores, 300
replications each. Every entry lies within Monte Carlo error of the
nominal $0.90$, including the never-trained network at $0.88$--$0.91$
and the network trained on permuted labels at $0.88$--$0.91$. Sabotaged
weights cost set length, their mean lengths being $52$--$88$ out of
$101$ candidates against $7$--$32$ for the trained scores, but never
validity:
this is Theorem~\ref{thm:coverage} made visible, and it is the property
that no post-hoc calibration of a fully learned detector provides for
free.

\begin{table}[t]
\centering
\caption{Empirical coverage of $\cset$ at nominal $1-\alpha=0.90$
(mean shift $\delta=2$, $n=120$, $\tau=60$, 300 replications; MC-SE
$\approx0.017$).}
\label{tab:coverage}
\small
\begin{tabular}{lcccc}
\toprule
Score & Gaussian & $t_3$ & Cauchy & Lognormal \\
\midrule
ARC learned combination $A^{w}$ & 0.917 & 0.963 & 0.907 & 0.903 \\
ARC max combination $A^{\mathrm{comb}}$ & 0.893 & 0.923 & 0.890 & 0.913 \\
ARC location channel & 0.910 & 0.917 & 0.900 & 0.890 \\
ARC network (trained) & 0.913 & 0.930 & 0.893 & 0.900 \\
ARC network (epoch 1) & 0.883 & 0.937 & 0.907 & 0.903 \\
ARC network (random weights) & 0.880 & 0.907 & 0.907 & 0.890 \\
ARC network (mistrained) & 0.880 & 0.907 & 0.907 & 0.880 \\
Raw CUSUM (plug-in) & 0.897 & 0.903 & 0.910 & 0.937 \\
Two-sample $t$ (plug-in LR) & 0.897 & 0.903 & 0.910 & 0.937 \\
\bottomrule
\end{tabular}
\end{table}

\subsection{Invariance: exactly identical sets}\label{sec:sim-invariance}

\begin{figure}[t]
\centering
\includegraphics[width=\textwidth]{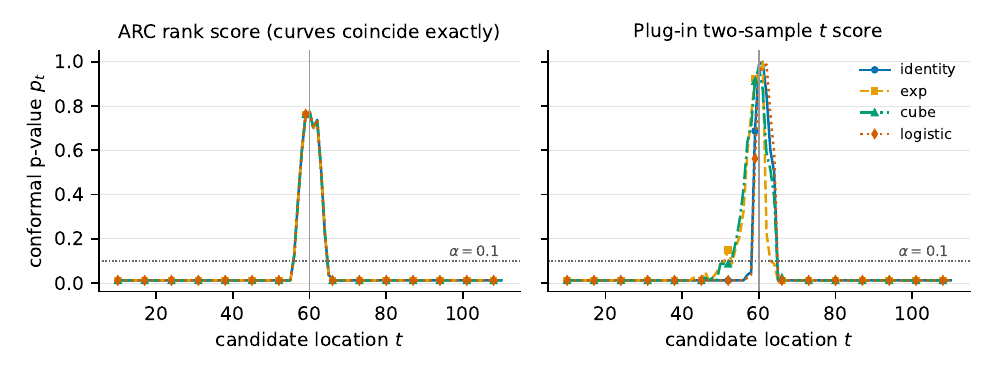}
\caption{Conformal p-value profiles $p_t$ on one Gaussian mean-shift
window ($n=120$, $\tau=60$, vertical line) and on its $\exp$, cube, and
logistic images, with shared permutation randomness. Left: the ARC
learned combination, for which the four curves coincide exactly
(Theorem~\ref{thm:invariance}), so one profile is visible. Right: the
plug-in two-sample $t$ score changes with each transform.}
\label{fig:invariance}
\end{figure}

Eighty Gaussian mean-shift windows ($\delta=2$) were each transformed by
three strictly increasing maps, namely $\exp(x)$, $x^3$, and the
logistic $1/(1+e^{-x})$, and the full procedure was re-run on each transformed
window with shared permutation randomness. For all eight rank-based
scores (two channels, two fixed combinations, four network weight
states) the maximal absolute p-value change across all candidates,
replications, and transforms was exactly $0$, and the confidence set was
identical to the untransformed one in $80/80$ replications: the
invariance of Theorem~\ref{thm:invariance} is not approximate.
Figure~\ref{fig:invariance} displays the mechanism on a single window:
the four transformed copies of the data produce four exactly coincident
ARC p-value profiles, whereas the plug-in profiles differ visibly,
since each transform presents the plug-in score with a different
estimation problem and elicits a different answer.
Table~\ref{tab:invariance} reports mean set lengths.

 The plug-in scores
are slightly \emph{shorter} at the Gaussian identity ($6.7$--$6.8$
vs.\ $7.8$ for the learned ARC combination), which is the classical
price of invariance of Remark~\ref{rem:price} made visible; under the
$\exp$ image of the very same data their sets inflate more than
threefold ($6.7\to21.6$ and $6.8\to22.1$) while every ARC length is
unchanged to the last replication.

\begin{table}[t]
\centering
\caption{Mean confidence-set length ($n=120$, $\alpha=0.10$, 80
replications) on the same windows under strictly increasing transforms.
Rank-based ARC scores are exactly invariant; Gaussian plug-in scores
retain coverage but inflate.}
\label{tab:invariance}
\small
\begin{tabular}{lcccc}
\toprule
Score & identity & $\exp$ & cube & logistic \\
\midrule
ARC learned combination $A^{w}$ & 7.83 & 7.83 & 7.83 & 7.83 \\
ARC max combination $A^{\mathrm{comb}}$ & 8.79 & 8.79 & 8.79 & 8.79 \\
ARC network (trained) & 15.20 & 15.20 & 15.20 & 15.20 \\
ARC network (random weights) & 61.46 & 61.46 & 61.46 & 61.46 \\
Raw CUSUM (plug-in) & 6.73 & 21.56 & 15.30 & 6.65 \\
Two-sample $t$ (plug-in LR) & 6.75 & 22.14 & 15.69 & 6.66 \\
\bottomrule
\end{tabular}
\end{table}

\subsection{Set length across distributions and change types}
\label{sec:sim-length}

\begin{figure}[t]
\centering
\includegraphics[width=\textwidth]{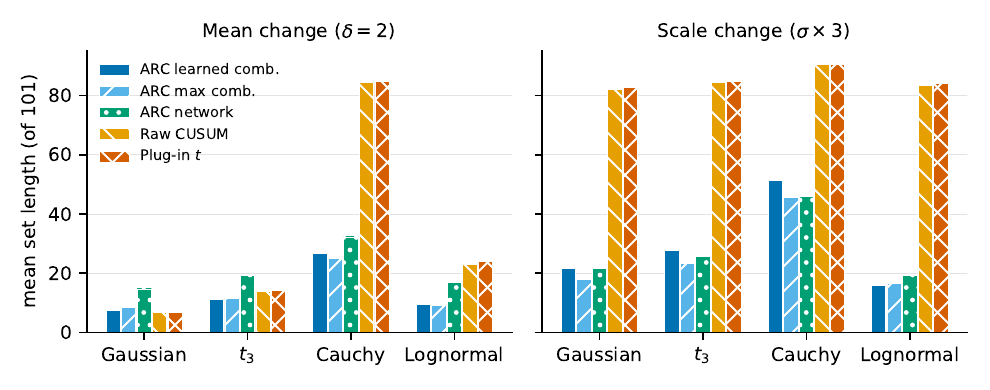}
\caption{Mean confidence-set length by marginal distribution and score
(values from Table~\ref{tab:length}). Coverage is at the nominal level
for every bar; length is where the scores separate. The plug-in scores
are marginally shorter at the Gaussian and collapse to near-vacuous
sets under the Cauchy, whereas the ARC scores deteriorate smoothly and,
by Theorem~\ref{thm:invariance}, their bars would be exactly identical
under any monotone re-expression of the data.}
\label{fig:lengths}
\end{figure}

Table~\ref{tab:length} and Figure~\ref{fig:lengths} report mean set
lengths for mean and scale changes. Three conclusions follow. First,
the efficiency ordering at the Gaussian is the classical one: the
plug-in scores are shortest at $6.8$, and the learned ARC combination
pays approximately ten percent at $7.5$, which is the numerical
counterpart of Remark~\ref{rem:price}. Second, the ordering reverses as
soon as the marginal departs from the Gaussian: at $t_3$ the learned
combination is already shorter, $11.2$ against $14.1$; at the lognormal
it is $2.6$ times shorter, $9.3$ against $23.9$; and at the Cauchy the
plug-in sets are effectively vacuous at $84.9$ of $101$ candidates,
while the learned combination reports $26.6$ and the location channel
$21.8$. Coverage is universal, whereas length is where the plug-in
scores deteriorate. Third, under scale changes the mean-oriented
plug-in scores are uninformative by construction, at $82$--$91$, while
the ARC scale channel and max combination localize at $15.8$--$45.5$.
This is a statement about the type coverage of the score bank rather
than about the plug-in family as such, since a variance-oriented
plug-in would face the same heavy-tail fragility.

\begin{table}[t]
\centering
\caption{Mean set length ($\alpha=0.10$, $n=120$; mean cells 300 reps,
scale cells 250 reps). Coverage for every entry is at, or within Monte
Carlo error of, the nominal $0.90$; the smallest observed value
($0.852$, $t_3$ scale, 250 reps) was rechecked at $1{,}000$
replications and returned $0.885$--$0.915$ across all scores
(MC-SE $\approx 0.010$; result file
\texttt{rev\_t3scale\_recheck.json}), confirming Monte Carlo
fluctuation rather than a coverage deficit.}
\label{tab:length}
\small
\begin{tabular}{lcccc@{\qquad}cccc}
\toprule
& \multicolumn{4}{c}{Mean change ($\delta=2$)} &
\multicolumn{4}{c}{Scale change ($\sigma\times3$)} \\
\cmidrule(r){2-5}\cmidrule(l){6-9}
Score & Gau. & $t_3$ & Cau. & Logn. & Gau. & $t_3$ & Cau. & Logn. \\
\midrule
ARC learned comb.\ $A^{w}$ & 7.5 & 11.2 & 26.6 & 9.3 & 21.8 & 27.8 & 51.4 & 15.8 \\
ARC max comb.\ $A^{\mathrm{comb}}$ & 8.3 & 11.6 & 24.9 & 9.2 & 18.0 & 23.3 & 45.5 & 16.4 \\
ARC location channel & 8.2 & 10.7 & 21.8 & 9.9 & 87.5 & 86.5 & 88.3 & 51.1 \\
ARC scale channel & 53.9 & 71.4 & 82.5 & 46.5 & 15.8 & 20.3 & 40.7 & 16.8 \\
ARC network (trained) & 14.7 & 19.4 & 32.2 & 16.9 & 21.6 & 25.6 & 45.8 & 19.4 \\
Raw CUSUM (plug-in) & 6.8 & 13.8 & 84.5 & 23.1 & 82.0 & 84.3 & 90.5 & 83.4 \\
Two-sample $t$ (plug-in LR) & 6.8 & 14.1 & 84.9 & 23.9 & 82.8 & 84.8 & 90.6 & 84.1 \\
\bottomrule
\end{tabular}
\end{table}

\subsection{Transfer: Gaussian-trained scores on heavy tails}
\label{sec:sim-transfer}

The transfer reading of Table~\ref{tab:length} isolates the two learned
scores, both fitted on Gaussian synthetic data only and frozen. Moving
from Gaussian to Cauchy test data multiplies the plug-in $t$-score's
mean length by $12.5$ ($6.8\to84.9$); the frozen ARC network grows by a
factor of $2.2$ ($14.7\to32.2$) and the learned rank combination by
$3.5$ ($7.5\to26.6$). For the ARC scores this growth is entirely a
property of the rank experiment $(F_0,F_1)$ rather than of the training
mismatch: by Theorem~\ref{thm:invariance} the same lengths would be
obtained for any monotone re-expression of the test data, whereas
the plug-in score would produce yet another, unpredictable length on
each. Coverage stays nominal for every score throughout
(Table~\ref{tab:coverage}).

\subsection{Type-aware ablation}\label{sec:sim-ablation}

Comparing rows of Table~\ref{tab:length} quantifies the price of not
knowing the change type. Under a mean change, the oracle (location
channel, $8.2$ at the Gaussian) is matched or beaten by both
combinations ($7.5$--$8.3$); under a scale change, the oracle scale
channel ($15.8$ at the Gaussian) is approached by the max combination
($18.0$) and the balanced learned combination ($21.8$). The oracle gap
of the max combination stays below a factor of $1.15$ across every mean
and scale cell, and at the lognormal it even undercuts the
single-channel oracle. Not knowing the change type therefore costs at
most fifteen percent in length, whereas selecting the wrong single
channel can cost a factor of five or more, as the location channel does
under scale changes at $51$--$88$.

\subsection{Conditional-on-detection reading}\label{sec:sim-cond}

Section~\ref{sec:pipeline} reports the confidence set unconditionally,
and a natural second-round question is how much the coverage statement
erodes for a user who only reads the set after the existence test
fires. We measured this directly (result file
\texttt{rev\_conditional.json}): with the existence test taken as the
whole-window permutation test of the combined score ($B=199$,
$\alpha=0.10$) and $300$ replications per cell, the Gaussian mean-shift
design at $\delta=2$ detects in $100\%$ of replications, so
conditional and marginal coverage coincide at $0.900$; at the weaker
$\delta=0.8$ the detection rate is $0.973$ and coverage conditional on
detection is $0.921$ against a marginal $0.920$ (mean lengths $39.5$
conditional vs.\ $40.9$ marginal). At these signal strengths the
selection effect is negligible because detection is nearly certain
whenever localization is informative; in severely underpowered regimes
the conditional guarantee genuinely requires the post-detection
machinery of \citet{saha2026post,saha2026distributionfree}, as stated
in Section~\ref{sec:pipeline}.

\subsection{Dependence preview}

Under AR(1) dependence ($\phi=0.5$) the i.i.d.-permutation procedure
undercovers catastrophically (coverage $0.30$ for the learned
combination at nominal $0.90$); circular block permutations with block
length $10$ restore $0.84$--$0.86$ at a substantial length cost
($2.6\to16.7$). Exactness is genuinely lost, not merely dented, and the
block variant is a repair, not a cure; Appendix~\ref{app:ar1} tabulates
all scores.

\section{Real Data: Well-Log Localization}\label{sec:real}

We analyse the classic well-log series, a record of nuclear magnetic response
measured while drilling, a standard changepoint benchmark whose level
shifts are contaminated by outlier bursts
\citep{oruanaidh1996numerical,fearnhead2019changepoint}, in the
version curated by the Turing Change Point Dataset (TCPD;
675 observations), which provides independent changepoint annotations from
five human annotators \citep{vandenburg2020evaluation}.

\paragraph{Protocol.} We call an index a \emph{consensus} changepoint
if at least three of the five annotators marked it (nine indices:
$179$, $255$, $281$, $311$, $343$, $402$, $412$, $422$, $432$). The
paper's model assumes a single changepoint per window, so we form one
window of length $48$ centred on each consensus changepoint whose
nearest consensus neighbour is farther than half a window away; this
yields five windows (around $179$, $255$, $281$, $311$, $343$) and
excludes the cluster $402$--$432$, whose spacings of about ten
observations place several changes in any reasonable window. Within
each window the annotated location is candidate $24$ of the
$33$ candidates ($m_0=8$, $m_1=4$, $B=99$, $\alpha=0.10$); the
network score and combination weights are those of
Section~\ref{sec:sim}, trained on synthetic Gaussian data and then
frozen, so that nothing is tuned on the well-log.

\begin{table}[t]
\centering
\caption{Well-log: 90\% localization sets per window. Each entry is the
set length out of 33 candidates, with \checkmark\ when the set contains
the consensus annotation. The empty sets in the first window are a
model-misfit alarm, discussed in the text.}
\label{tab:welllog}
\small
\begin{tabular}{lcccccc}
\toprule
Annotation & ARC learned & ARC max & Location & Scale & ARC network &
Plug-in $t$ \\
\midrule
179 & 0 & 0 & 2 & 16 \checkmark & 4 & 2 \\
255 & 4 \checkmark & 7 \checkmark & 6 \checkmark & 17 \checkmark & 8 \checkmark & 7 \checkmark \\
281 & 5 \checkmark & 5 \checkmark & 7 \checkmark & 12 \checkmark & 8 \checkmark & 3 \checkmark \\
311 & 4 \checkmark & 6 \checkmark & 5 \checkmark & 20 \checkmark & 6 \checkmark & 4 \checkmark \\
343 & 3 \checkmark & 2 \checkmark & 3 & 13 \checkmark & 6 \checkmark & 3 \\
\bottomrule
\end{tabular}
\end{table}

\begin{figure}[t]
\centering
\includegraphics[width=\textwidth]{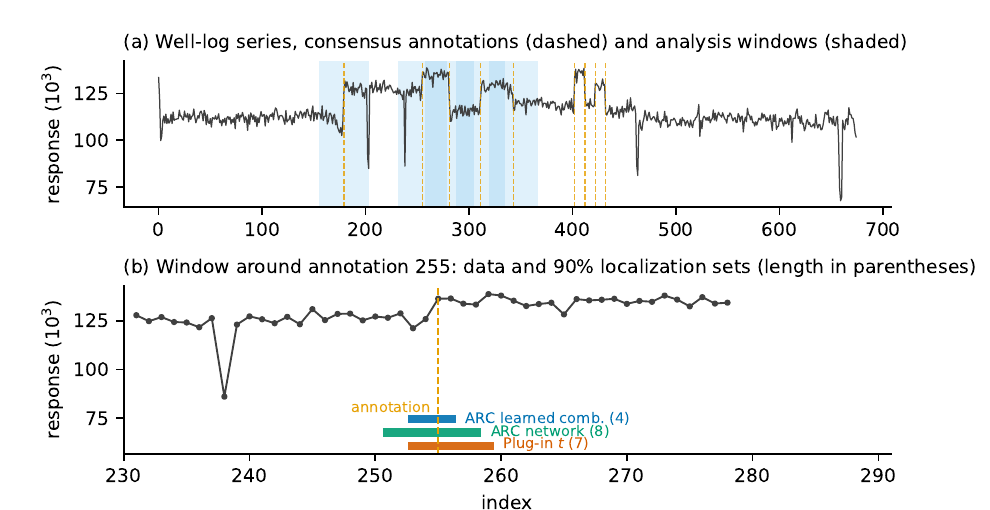}
\caption{Well-log analysis. (a) The full series with the nine consensus
annotations (dashed) and the five single-changepoint analysis windows
(shaded). (b) The window around annotation 255: data, annotation, and
the 90\% localization sets of three scores (bar length = set extent;
set size in parentheses).}
\label{fig:welllog}
\end{figure}

\paragraph{Results.} Table~\ref{tab:welllog} and
Figure~\ref{fig:welllog} summarize the outcome. In four of the five
windows the ARC learned combination returns sets of three to five
candidates that contain the annotation, which constitutes an
actionable uncertainty statement of $\pm2$ observations at 90\%
confidence under no distributional assumption. The plug-in $t$ sets are
of comparable length here, consistent with the Gaussian column of
Table~\ref{tab:length}, since within these windows the marginals are
locally near-Gaussian and the advantage established by
Theorem~\ref{thm:invariance} functions as insurance rather than as a
guaranteed improvement. At annotation $343$ the plug-in and pure
location sets, locally $\{21,22,23\}$, miss the annotated index by one
position, which lies within the $\pm1$ ambiguity of the annotation
convention, whereas the combined ARC sets contain it.

The window around $179$ is instructive: the combined scores return
empty sets. An empty conformal set is not a failure mode but a
certificate, since under the assumed single-changepoint
piecewise-i.i.d.\ model it occurs with probability at most $\alpha$,
and it therefore signals model misfit in this window at level $0.10$.
The signal is corroborated externally, as this is precisely the region
where the human annotators disagree, two placing the change at $177$
and three at $179$; the location and scale channels point at different
candidates (Table~\ref{tab:welllog}), and the transition there is
gradual rather than abrupt. A procedure that must always return an
interval would have silently reported a wrong one.

\section{Discussion}\label{sec:discussion}

Five limitations bound the claims. First, the theory is for a single
changepoint in an offline window; multiple changepoints can be handled in
practice by seeded or wild binary segmentation wrappers
\citep{fryzlewicz2014wild,baranowski2019not} around ARC windows, but the
simultaneous coverage of the resulting sets is not established here, and
the multiscale machinery of \citet{cui2026art} is the natural bridge.
Second, exact coverage requires within-segment independence; under serial
dependence the block variant restores validity only approximately
(Appendix~B), and a rank theory for dependent data
\citep{banerjee2026markov} is the direction we consider most promising.
Third, our sets are marginal, not conditional on a detection event;
conditional coverage after an alarm is the harder target addressed by
\citet{saha2026post,saha2026distributionfree}. Fourth, type adaptivity is
bought by freezing the combination rule in advance, and the oracle gap
in Section~\ref{sec:sim-ablation}, while modest, is not zero. Fifth,
the piecewise-exchangeable model excludes trend-type changes
(Remark~\ref{rem:scope}); a rigorous treatment of drifting segments
within a conformal localization framework is open.

Two extensions follow naturally. Online ARC would combine the rank
channels with e-value or anytime-valid machinery
\citep{wei2026online,lee2026fdr}, whose exchangeability-martingale
roots go back to \citet{vovk2003testing}. In the multivariate direction,
coordinate attribution with error control
\citep{ouerfelli2026posthoc,lan2026wave} could be conformalized with
rank scores per coordinate, extending type-awareness to ``which
coordinate changed'', a step toward attribution-aware localization on
object- and distribution-valued streams
\citep{zhang2026object,zeng2026beyond}.

\section{Conclusion}\label{sec:conclusion}

Coverage of conformal changepoint localization is universal; efficiency
is not. ARC makes efficiency the object of design. Scoring the window
through within-segment ranks, and freezing every learned component
before the test window is seen, yields confidence sets that are
finite-sample valid for any weight configuration and almost surely
invariant to monotone distortions of the data. Validity is universal,
whereas ranks are what render efficiency transferable.

\section*{Declarations}

\paragraph{Data availability.} All results in this paper are fully
reproducible from the accompanying code (\texttt{arc\_core.py},
\texttt{run\_experiments.py}, \texttt{real\_data\_welllog.py},
\texttt{rev\_experiments.py}; fixed
seeds) and the generated result files, both of which accompany this
preprint as ancillary files. The implementation uses Python with NumPy
only (no GPU); the full study, network training included, runs in about
one hour on a laptop CPU, and dependency versions are pinned in the
accompanying \texttt{requirements.txt}. The well-log series and its annotations are publicly
available from the Turing Change Point Dataset repository
\citep{vandenburg2020evaluation}.


\paragraph{Conflict of interest.} The authors declare no conflict of
interest.

\paragraph{Ethics.} This study uses synthetic data and a public,
fully de-identified geophysical benchmark series (well-log, distributed
with the Turing Change Point Dataset); no human-subject data are
involved.

\paragraph{AI usage disclosure.} Large-language-model assistance was
used for literature triage, drafting, and code scaffolding under author
direction; all experiments were executed and verified computationally,
all numbers in the paper are read from the generated result files, and
the authors take full responsibility for the content.

\appendix

\section{Proofs}\label{app:proofs}

\subsection{Proof of Theorem~\ref{thm:coverage}}
Under $H_{0,\tau}$ the left and right segments are independent samples of
exchangeable (i.i.d.) variables. Conditional on the two segment
multisets, the observed window is uniformly distributed on the orbit of
within-segment permutations, and $W_\tau$ is a deterministic function of
the window and the frozen parameters (which are independent of the test
window by assumption). The Monte Carlo permutation p-value with the
$+1$-correction in \eqref{eq:pval} is therefore super-uniform
\citep{hemerik2018exact,lehmann2005testing}:
$\pr(p_\tau\le\alpha)\le\alpha$, whence
$\pr(\tau\in\cset)=\pr(p_\tau>\alpha)\ge1-\alpha$. No property of the
training procedure enters the argument. \qed

\subsection{Proof of Lemma~\ref{lem:ranks} and
Theorem~\ref{thm:invariance}}
Strict monotonicity preserves all pairwise order relations, so the rank
vector of every segment, and of every permuted segment, is unchanged by
$g$; ARC statistics are deterministic functions of these rank vectors,
so every $W_t$ and $W_t^{(b)}$ is unchanged, and with shared permutation
draws every $p_t$ in \eqref{eq:pval} is unchanged, hence
$\cset(g(X))=\cset(X)$ almost surely. For the transfer statement,
observe that if $X$ follows $(F_0,F_1,\tau)$ then $g(X)$ follows
$(F_0\circ g^{-1},F_1\circ g^{-1},\tau)$ and, by the coupling just
established, $\cset(g(X))=\cset(X)$; thus any functional of
$(\cset,\tau)$ has the same law under both models. Since every pair
$(G_0,G_1)$ in the monotone orbit of $(F_0,F_1)$ arises this way, the
law is constant on the orbit. \qed

\subsection{Proof sketch of Proposition~\ref{prop:length}}
Fix $\varepsilon>0$ and a candidate $t$ with $t\le\tau-\varepsilon n$
(the mirrored case is symmetric). The right segment $X_{t+1:n}$ then
contains the true changepoint at relative position bounded away from its
edges, and splits it into subsamples of sizes proportional to $n$ on
both sides. The internal Wilcoxon evidence at the true split is a
two-sample $U$-statistic with drift
$|\rho-1/2|\sqrt{s(m-s)\,12/(m+1)}\asymp\sqrt n\,|\rho-1/2|$; Hoeffding's
inequality for $U$-statistics gives
$A^{\mathrm{loc}}\ge c\sqrt n\,|\rho-1/2|$ with probability
$1-e^{-c'n}$. Under within-segment permutation the same statistic is a
maximum of $O(n)$ standardized rank CUSUMs with sub-Gaussian tails, so
the $(1-\alpha)$ permutation quantile is $O(\sqrt{\log n})$; a union
bound over the $B$ draws and over $t$ yields
$\max_{|t-\tau|\ge\varepsilon n}p_t\le\alpha$ with probability tending
to one exponentially. Hence
$\cset\subseteq(\tau-\varepsilon n,\tau+\varepsilon n)$ eventually, for
every $\varepsilon>0$. Two steps are stated here without full proof and
would need to be discharged in a rigorous account: the sub-Gaussian tail
bound for the maximum of standardized rank CUSUMs under permutation
(available in principle from H\'ajek-projection tail inequalities for
linear rank statistics), and the union bound over the $B$ Monte Carlo
draws, which requires $B$ to grow at least logarithmically in $n$.
Constants are not optimized; see
\citet{hore2026conformal,bhattacharya2025length,madrid2021optimal} for
sharper analyses under stronger assumptions. \qed

\section{Dependence: AR(1) study}\label{app:ar1}

Windows follow a stationary AR(1) process with $\phi=0.5$, unit
marginal variance, and a mean shift $\delta=2$ at $\tau=60$
($n=120$, 250 replications, $\alpha=0.10$). Table~\ref{tab:ar1}
compares i.i.d.\ within-segment permutations with circular block
permutations (block length $10$). Positive autocorrelation makes
locally smooth excursions that every internal-heterogeneity statistic
reads as changepoint evidence, so the i.i.d.-permutation p-value
rejects even at the true split: sharp scores collapse to
$0.24$--$0.30$ coverage. Block permutations restore $0.84$--$0.92$,
still below nominal because the block length preserves the dependence
structure only approximately, at three-to-seven-fold longer sets. All
exactness claims in the main text are therefore confined to
within-segment independence, as stated.

\begin{table}[h]
\centering
\caption{AR(1) dependence ($\phi=0.5$): coverage / mean length at
nominal $0.90$.}
\label{tab:ar1}
\small
\begin{tabular}{lcc@{\qquad}cc}
\toprule
& \multicolumn{2}{c}{i.i.d.\ permutation} &
\multicolumn{2}{c}{Block permutation ($L=10$)} \\
\cmidrule(r){2-3}\cmidrule(l){4-5}
Score & coverage & length & coverage & length \\
\midrule
ARC learned combination & 0.296 & 2.6 & 0.844 & 16.7 \\
ARC max combination & 0.296 & 3.1 & 0.856 & 16.8 \\
ARC location channel & 0.248 & 2.7 & 0.836 & 16.7 \\
ARC scale channel & 0.776 & 34.1 & 0.908 & 74.0 \\
ARC network (trained) & 0.552 & 9.1 & 0.868 & 28.9 \\
Raw CUSUM (plug-in) & 0.240 & 2.2 & 0.864 & 13.8 \\
Two-sample $t$ (plug-in LR) & 0.240 & 2.2 & 0.856 & 14.3 \\
\bottomrule
\end{tabular}
\end{table}

\section{Sensitivity to the number of permutations}\label{app:bsens}

The Monte Carlo p-values \eqref{eq:pval} lie on a grid of spacing
$1/(B+1)$, so small $B$ makes the procedure conservative by at most one
grid step and adds Monte Carlo noise to every $p_t$. To verify that the
paper's conclusions are not driven by the default $B=79$, the Gaussian
mean-shift cell ($\delta=2$, $n=120$, $200$ replications, MC-SE
$\approx0.021$) was rerun at $B\in\{79,199,999\}$
(result file \texttt{rev\_b\_sensitivity.json}).
Table~\ref{tab:bsens} shows that empirical coverage stays within Monte
Carlo error of the nominal $0.90$ at every $B$ and, more to the point,
that mean set lengths are flat in $B$ for every score (the learned
combination moves from $7.4$ to $7.5$; the plug-in $t$ from $6.5$ to
$6.6$): the efficiency comparisons of Section~\ref{sec:sim} are not an
artefact of p-value discretization.

\begin{table}[h]
\centering
\caption{$B$-sensitivity (Gaussian mean shift, $\delta=2$,
$\alpha=0.10$, 200 replications): coverage / mean length.}
\label{tab:bsens}
\small
\begin{tabular}{lccc}
\toprule
Score & $B=79$ & $B=199$ & $B=999$ \\
\midrule
ARC learned combination & 0.880 / 7.4 & 0.930 / 7.5 & 0.915 / 7.5 \\
ARC max combination & 0.895 / 8.1 & 0.910 / 8.1 & 0.920 / 8.0 \\
ARC location channel & 0.870 / 7.9 & 0.935 / 8.1 & 0.930 / 7.8 \\
ARC network (trained) & 0.875 / 14.1 & 0.925 / 14.3 & 0.915 / 13.5 \\
Two-sample $t$ (plug-in LR) & 0.895 / 6.5 & 0.915 / 6.6 & 0.930 / 6.6 \\
\bottomrule
\end{tabular}
\end{table}

\bibliographystyle{plainnat}
\bibliography{arc}

\end{document}